%% file: main.tex
\documentclass[10pt,twocolumn,letterpaper]{article}
\pdfoutput=1
\usepackage{cvpr}
\usepackage{times}
\usepackage{epsfig}
\usepackage{graphicx}
\usepackage{amsmath}
\usepackage{booktabs}
\usepackage{amssymb}
\DeclareMathOperator*{\argmin}{arg\min}
\graphicspath{{gfx/}}

\usepackage[pagebackref=true,breaklinks=true,letterpaper=true,colorlinks,bookmarks=false]{hyperref}

\cvprfinalcopy %

\def\cvprPaperID{25} %

\ifcvprfinal\fi
\begin{document}

\title{3D Face Morphable Models ``In-the-Wild''}
\author{James Booth
\and
Epameinondas Antonakos
\and
Stylianos  Ploumpis
\and
George  Trigeorgis
\and
Yannis  Panagakis
\and
Stefanos Zafeiriou
\and
Imperial College London, UK
\\
{\tt\small \{james.booth,e.antonakos,s.ploumpis,g.trigeorgis,i.panagakis,s.zafeiriou\}@imperial.ac.uk}
}

\maketitle
\begin{abstract}
3D Morphable Models (3DMMs) are powerful statistical models of 3D facial shape and texture, and among the state-of-the-art methods for reconstructing facial shape from single images.
With the advent of new 3D sensors, many 3D facial datasets have been collected containing both neutral as well as expressive faces.
However, all datasets are captured under controlled conditions.
Thus, even though powerful 3D facial shape models can be learnt from such data, it is difficult to build statistical texture models that are sufficient to reconstruct faces captured in unconstrained conditions (``in-the-wild'').
In this paper, we propose the first, to the best of our knowledge, ``in-the-wild'' 3DMM by combining a powerful statistical model of facial shape, which describes both identity and expression, with an ``in-the-wild'' texture model.
We show that the employment of such an ``in-the-wild'' texture model greatly simplifies the fitting procedure, because there is no need to optimize with regards to the illumination parameters.
Furthermore, we propose a new fast algorithm for fitting the 3DMM in arbitrary images.
Finally, we have captured the first 3D facial database with relatively unconstrained conditions and report quantitative evaluations with state-of-the-art performance.
Complementary qualitative reconstruction results are demonstrated on standard ``in-the-wild'' facial databases.
An open source implementation of our technique is released as part of the Menpo Project~\cite{menpo14}.
\end{abstract}

\input{introduction}
\input{training}
\input{fitting}
\input{dataset}
\input{experiments}
\input{conclusion}

{\small
\bibliographystyle{ieee}
\bibliography{egbib}
}

\end{document}

%% file: introduction.tex
\begin{figure}
    \centering
    \includegraphics[width=\linewidth]{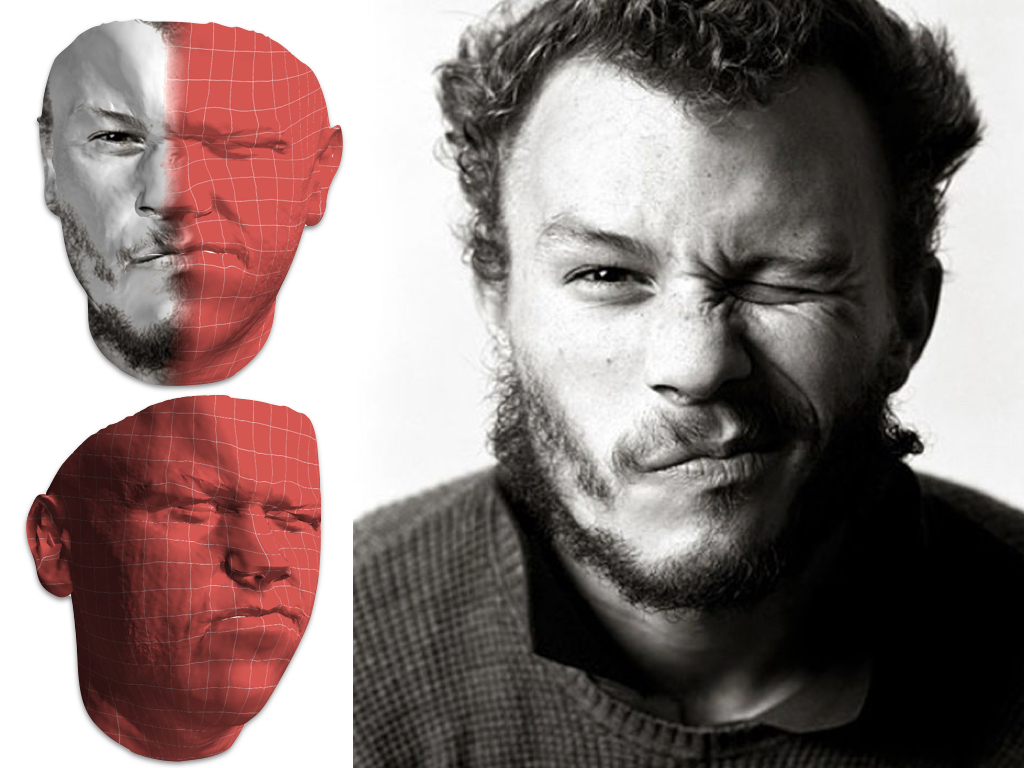}
    \caption{Our ``in-the-wild'' Morphable Model is capable of recovering accurate 3D facial shape for a wide variety of images.}
\label{fig:hero}
\end{figure}

\section{Introduction}
During the past few years, we have witnessed significant improvements in various
face analysis tasks such as face detection~\cite{fddbTech,zafeiriou2015survey}
and 2D facial landmark localization on static
images~\cite{xiong2013supervised,kazemi2014one,asthana2014incremental,tzimiropoulos2014gauss,zhu2015face,antonakos2015feature,antonakos2015active,trigeorgis2016mnemonic}.
This is primarily attributed to the fact that the community has made a
considerable effort to collect and annotate facial images
captured under unconstrained conditions~\cite{le2012interactive,zhu2012face,LFPW_belhumeur2013localizing,sagonas2013300,sagonas2016faces} (commonly referred to as ``in-the-wild'') and to the discriminative methodologies that can capitalise on the availability of
such large amount of data. Nevertheless, discriminative techniques cannot be
applied for 3D facial shape estimation ``in-the-wild'', due to lack of
ground-truth data.

3D facial shape estimation from single images has attracted the attention of
many researchers the past twenty years. The two main lines of research are \emph{(i)}~fitting a 3D Morphable Model
(3DMM)~\cite{blanz1999morphable,blanz2003face} and \emph{(ii)}~applying Shape from Shading (SfS) techniques~\cite{snape2015automatic,snape2014kernel,kemelmacher2013internet}.
The 3DMM fitting proposed in the work of
Blanz and Vetter~\cite{blanz1999morphable,blanz2003face} was among the first model-based 3D
facial recovery approaches. The method requires the construction of a 3DMM which
is a statistical model of facial texture and shape in a space where there are
explicit correspondences. The first 3DMM was built using 200 faces captured
in well-controlled conditions displaying only the neutral expression.
That is the reason why the method was only shown to work on real-world, but not ``in-the-wild'', images.
State-of-the-art SfS techniques capitalise on special multi-linear decompositions
that find an approximate spherical harmonic decomposition of the illumination.
Furthermore, in order to benefit from the large availability of ``in-the-wild'' images, these methods jointly reconstruct large collections of images. Nevertheless, even thought the results of \cite{snape2015automatic,kemelmacher2013internet} are quite interesting, given that there is no prior of the facial surface, the methods only recover 2.5D representations of the faces and particular smooth approximations of the facial normals.

3D facial shape recovery from a single image under ``in-the-wild'' conditions is still an open and challenging problem in computer vision mainly due to the fact that:
\begin{itemize}
\item The general problem of extracting the 3D facial shape from a single image
is an ill-posed problem which is notoriously difficult to be solved without the
use of any statistical priors for the shape and texture of faces. That is,
without prior knowledge regarding the shape of the object at-hand there are
inherent ambiguities present in the problem. The pixel intensity at a location
in an image is the result of a complex combination of the underlying shape of
the object, the surface albedo and normal characteristics, camera parameters
and the arrangement of scene lighting and other objects in the scene.
Hence, there are potentially infinite solutions to the problem.

\item Learning statistical priors of the 3D facial shape and texture for
``in-the-wild'' images is currently very difficult by using modern acquisition devices.
That is, even though there is a considerable improvement in 3D acquisition
devices, they still cannot operate in arbitrary conditions. Hence, all the
current 3D facial databases have been captured in controlled conditions.
\end{itemize}

With the available 3D facial data, it is feasible to learn a powerful statistical model of the facial shape that generalises well for both identity and expression \cite{cao2014facewarehouse,paysan20093d,booth3d}.
However, it is not possible to construct a statistical model of the facial texture that
generalises well for ``in-the-wild'' images and is, at the same time, in
correspondence with the statistical shape model.
That is the reason why current state-of-the-art 3D face reconstruction
methodologies rely solely on fitting a statistical 3D facial shape prior on a
sparse set of landmarks \cite{aldrian2013inverse,huber2015fitting}.

In this paper, we make a number of contributions that enable the use of 3DMMs for ``in-the-wild'' face reconstruction (Fig.~\ref{fig:hero}). In particular, our contributions are:

\begin{itemize}

\item We propose a methodology for learning a statistical texture model from ``in-the-wild'' facial images, which
is in full correspondence with a statistical shape prior that exhibits both
identity and expression variations.
Motivated by the success of feature-based
(e.g., HOG~\cite{dalal2005histograms}, SIFT~\cite{lowe1999object})
Active Appearance Models (AAMs)~\cite{antonakos2014hog,antonakos2015feature}
we further show how to learn feature-based texture models for 3DMMs.
We show that the advantage of using the ``in-the-wild'' feature-based texture model is that the fitting strategy gets simplified since there is not need to optimize with respect to the illumination parameters.

\item By capitalising on the recent advancements in fitting statistical
deformable models~\cite{papandreou2008adaptive,tzimiropoulos2013optimization,antonakos2015feature,alabort2016unified},
we propose a novel and fast algorithm for fitting ``in-the-wild'' 3DMMs. Furthermore, we make the
implementation of our algorithm publicly available, which we believe can be of
great benefit to the community, given the lack of robust open-source implementations for fitting 3DMMs.

\item Due to lack of ground-truth data, the majority of the 3D face
reconstruction papers report only qualitative results. In this paper, in order
to provide quantitative evaluations, we collected a new dataset of 3D facial surfaces, using Kinect Fusion~\cite{izadi2011kinectfusion,newcombe2011kinectfusion}, which has many ``in-the-wild'' characteristics, even though it is captured indoors.

\item We release an open source implementation of our technique as part of the Menpo Project.~\cite{menpo14}

\end{itemize}

The remainder of the paper is structured as follows.
In Section~\ref{sec:training} we elaborate on the construction of our ``in-the-wild'' 3DMM,
whilst in Section~\ref{sec:fitting} we outline the proposed optimization for fitting
``in-the-wild'' images with our model. Section~\ref{sec:dataset} describes our
new dataset, the first of its kind, to provide images with a ground-truth 3D facial shape that exhibit many ``in-the-wild'' characteristics. We outline a series of quantitative and qualitative experiments in Section~\ref{sec:experiments}, and end with conclusions in Section~\ref{sec:conclusion}.

%% file: training.tex
\section{Model Training}
\label{sec:training}

A 3DMM consists of three parametric models: the \emph{shape},
\emph{camera} and \emph{texture} models.

\subsection{Shape Model}
Let us denote the 3D mesh (shape) of an object with $N$ vertexes as a
$3N\times 1$ vector
\begin{equation}
\mathbf{s} = {\left[\mathbf{x}_1^\mathsf{T}, \ldots, \mathbf{x}_N^\mathsf{T}\right]}^\mathsf{T} = {\left[x_1, y_1, z_1, \ldots, x_N, y_N, z_N\right]}^\mathsf{T}
\label{equ:3D_shape_vector}
\end{equation}
where $\mathbf{x}_i={\left[x_i, y_i, z_i\right]}^\mathsf{T}$ are the
object-centered Cartesian coordinates of the $i$-th vertex. A 3D shape model can be constructed by first bringing a set of 3D training meshes into dense correspondence so that each is described with the same number of vertexes and all samples have a shared semantic ordering.
The corresponded meshes, $\left\lbrace\mathbf{s}_i\right\rbrace$, are then brought into a shape space by applying Generalized Procrustes Analysis and then Principal Component Analysis (PCA) is performed which results in
$\left\lbrace\bar{\mathbf{s}}, \mathbf{U}_s\right\rbrace$, where
$\bar{\mathbf{s}}\in\mathbb{R}^{3N}$ is the mean shape vector and
$\mathbf{U}_s\in\mathbb{R}^{3N\times n_s}$ is the orthonormal basis after
keeping the first $n_s$ principal components. This model can be used to
generate novel 3D shape instances using the function
$\mathcal{S}: \mathbb{R}^{n_s} \rightarrow \mathbb{R}^{3N}$ as
\begin{equation}
\mathcal{S}(\mathbf{p}) = \bar{\mathbf{s}} + \mathbf{U}_s \mathbf{p}
\label{equ:shape_instance}
\end{equation}
where $\mathbf{p}={\left[p_1,\ldots,p_{n_s}\right]}^\mathsf{T}$ are the $n_s$ \emph{shape parameters}.

\begin{figure}[!t]
    \centering
    \includegraphics[width=\linewidth]{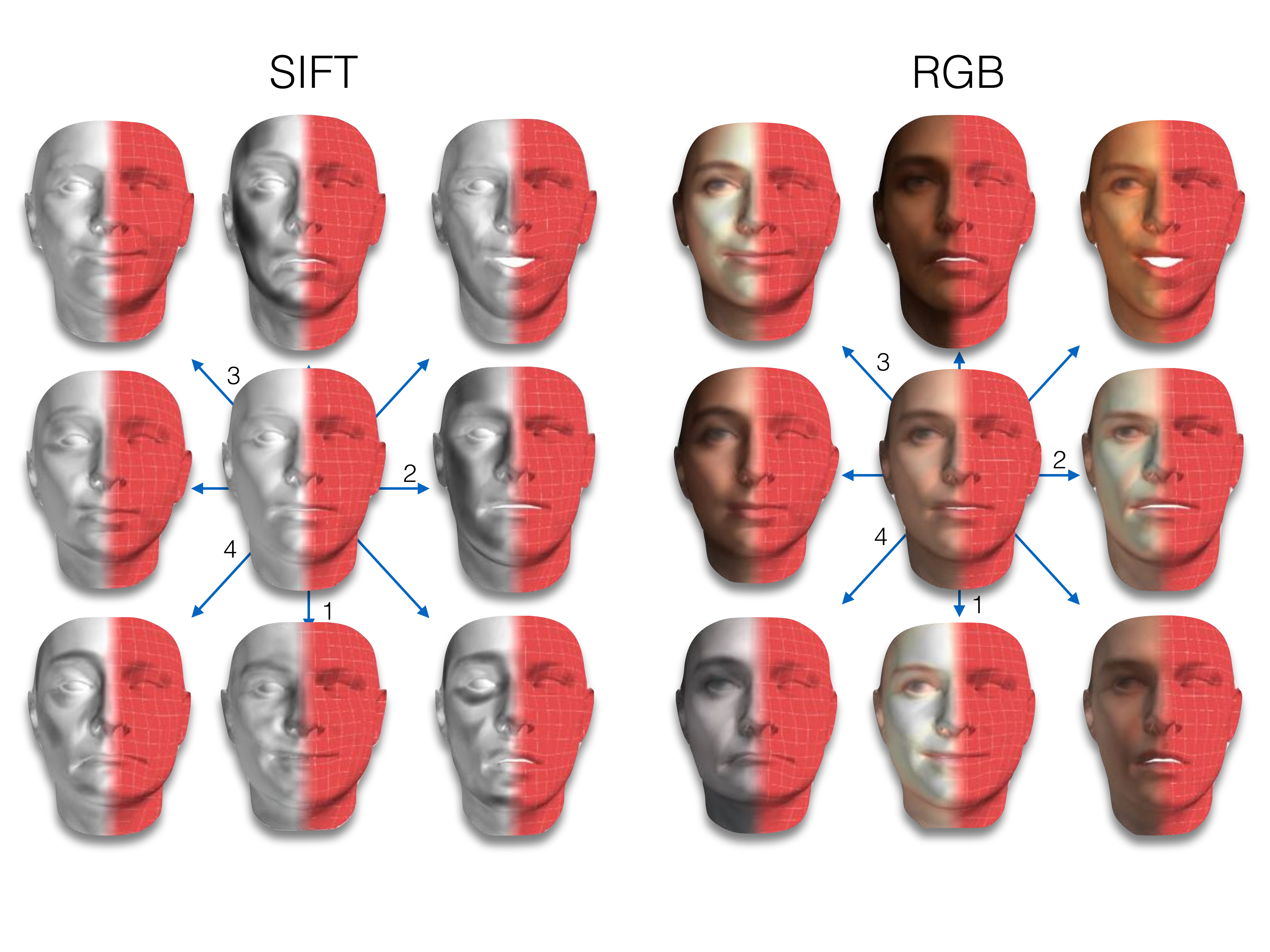}
    \caption{\emph{Left:} The mean and first four shape and SIFT texture principal components of our ``in-the-wild'' SIFT texture model. \emph{Right:} To aid in interpretation we also show the equivalent RGB basis.}
\label{fig:basis}
\end{figure}
\subsection{Camera Model}
The purpose of the camera model is to map (project) the object-centered Cartesian
coordinates of a 3D mesh instance $\mathbf{s}$ into 2D Cartesian coordinates on
an image plane. In this work, we employ a pinhole camera model, which utilizes a
perspective transformation. However, an orthographic projection model can also
be used in the same way.

\textbf{Perspective projection.} The projection of a 3D point
$\mathbf{x}={[x,y,z]}^\mathsf{T}$ into its 2D location in the image plane
$\mathbf{x}'={[x',y']}^\mathsf{T}$ involves two steps. First, the 3D point is
rotated and translated using a linear \emph{view transformation}, under the
assumption that the camera is still
\begin{equation}
{\left[v_x, v_y, v_z\right]}^\mathsf{T} = \mathbf{R}_v\mathbf{x} + \mathbf{t}_v
\label{equ:view_transform}
\end{equation}
where $\mathbf{R}_v\in\mathbb{R}^{3\times 3}$ and
$\mathbf{t}_v={[t_x,t_y,t_z]}^\mathsf{T}$ are the 3D rotation and translation
components, respectively. Then, a non-linear \emph{perspective transformation}
is applied as
\begin{equation}
\mathbf{x}' = \frac{f}{v_z}\left[\begin{array}{c}v_x\\v_y\end{array}\right] + \left[\begin{array}{c}c_x\\c_y\end{array}\right]
\label{equ:perspective_transform}
\end{equation}
where $f$ is the focal length in pixel units (we assume that the $x$ and $y$ components of the focal length are equal) and
${[c_x, c_y]}^\mathsf{T}$ is the principal point that is set to the image center.

\textbf{Quaternions.}
We parametrize the 3D rotation with quaternions~\cite{kuipers1999quaternions,wheeler1995iterative}.
The quaternion uses four parameters
$\mathbf{q}={\left[q_0,q_1,q_2,q_3\right]}^\mathsf{T}$ in order to express a 3D
rotation as
\begin{equation}
\mathbf{R}_v =
2\left[\begin{array}{ccc}
\frac{1}{2}-q_2^2-q_3^2 & q_1q_2-q_0q_3 & q_1q_3+q_0q_2\\
q_1q_2+q_0q_3 & \frac{1}{2}-q_1^2-q_3^2 & q_2q_3-q_0q_1\\
q_1q_3-q_0q_2 & q_2q_3+q_0q_1 & \frac{1}{2}-q_1^2-q_2^2
\end{array}\right]
\end{equation}
Note that by enforcing a unit norm constraint on the quaternion vector, i.e.
$\mathbf{q}^\mathsf{T}\mathbf{q}=1$, the rotation matrix constraints of
orthogonality with unit determinant are withheld. Given the unit norm property,
the quaternion can be seen as a three-parameter vector
${\left[q_1,q_2,q_3\right]}^\mathsf{T}$ and a scalar
$q_0=\sqrt{1-q_1^2-q_2^2-q_3^2}$. Most existing works on 3DMM parametrize the
rotation matrix $\mathbf{R}_v$ using the three Euler angles that define the
rotations around the horizontal, vertical and camera axes. Even thought Euler
angles are more naturally interpretable, they have strong disadvantages when
employed within an optimization procedure, most notably the solution ambiguity
and the gimbal lock effect. Parametrization based on quaternions overcomes these
disadvantages and further ensures computational efficiency, robustness and
simpler differentiation.

\textbf{Camera function.}
The projection operation performed by the camera model of the 3DMM can be
expressed with the function
$\mathcal{P}(\mathbf{s},\mathbf{c}): \mathbb{R}^{3N} \rightarrow \mathbb{R}^{2N}$,
which applies the transformations of Eqs.~\ref{equ:view_transform} and~\ref{equ:perspective_transform}
on the points of provided 3D mesh $\mathbf{s}$ with
\begin{equation}
\mathbf{c} = {\left[f, q_1, q_2, q_3, t_x, t_y, t_z\right]}^\mathsf{T}
\end{equation}
being the vector of \emph{camera parameters} with length $n_c=7$.
For abbreviation purposes, we represent the camera model of the 3DMM with the
function $\mathcal{W}: \mathbb{R}^{n_s,n_c} \rightarrow \mathbb{R}^{2N}$ as
\begin{equation}
\mathcal{W}(\mathbf{p},\mathbf{c})\equiv\mathcal{P}\left(\mathcal{S}(\mathbf{p}),\mathbf{c}\right)
\label{equ:camera_model}
\end{equation}
where $\mathcal{S}(\mathbf{p})$ is a 3D mesh instance using Eq.~\ref{equ:shape_instance}.

\subsection{``In-the-Wild'' Feature-Based Texture Model}
\label{sec:texture-model}
The generation of an ``in-the-wild'' texture model is a key component of the
proposed 3DMM. To this end, we take advantage of the existing large facial ``in-the-wild''
databases that are annotated in terms of sparse landmarks.
Assume that for a set of $M$ ``in-the-wild'' images
$\left\lbrace\mathbf{I}_i\right\rbrace_1^M$, we have access to the associated
camera and shape parameters $\left\lbrace\mathbf{p_i}, \mathbf{c_i}\right\rbrace$.
Let us also define a \emph{dense} feature extraction function
\begin{equation}
\mathcal{F}: \mathbb{R}^{H\times W}\rightarrow\mathbb{R}^{H\times W \times C}
\label{equ:features}
\end{equation}
where $C$ is the number of channels of the feature-based image. For each image,
we first compute its feature-based representation as
$\mathbf{F}_i = \mathcal{F}(\mathbf{I}_i)$ and then use Eq.~\ref{equ:camera_model}
to sample it at each vertex location to build back a vectorized texture sample
$\mathbf{t}_i = \mathbf{F}_i\left(\mathcal{W}(\mathbf{p_i},\mathbf{c_i})\right) \in \mathbb{R}^{CN}$.
This texture sample will be nonsensical for some regions mainly due to self-occlusions
present in the mesh projected in the image space
$\mathcal{W}(\mathbf{p_i},\mathbf{c_i})$.
To alleviate these issues, we cast a ray from the camera to each vertex and test
for self-intersections with the triangulation of the mesh in order to learn a
per-vertex occlusion mask $\mathbf{m}_i\in\mathbb{R}^{N}$ for the projected sample.

Let us create the matrix
$\mathbf{X}=\left[\mathbf{t}_1, \ldots, \mathbf{t}_M\right]\in\mathbb{R}^{CN \times M}$
by concatenating the $M$ grossly corrupted feature-based texture vectors with missing entries that are represented by the masks $\mathbf{m}_i$.
To robustly build a texture model based on this heavily contaminated incomplete
data, we need to recover a low-rank matrix $\mathbf{L}\in\mathbb{R}^{CN \times M}$
representing the clean facial texture and a sparse matrix $\mathbf{E}\in\mathbb{R}^{CN \times M}$
accounting for gross but sparse non-Gaussian noise such that $\mathbf{X}=\mathbf{L}+\mathbf{E}$.
To simultaneously recover both $\mathbf{L}$ and $\mathbf{E}$ from incomplete and
grossly corrupted observations, the Principal Component Pursuit with missing
values~\cite{shang2014robust} is solved
\begin{equation}
\begin{aligned}
\argmin_{\mathbf{L},\mathbf{E}} &~\lVert\mathbf{L}\rVert_* + \lambda \lVert\mathbf{E}\rVert_1\\
\mbox{s.t.} &~\mathcal{P}_{\Omega}( \mathbf{X}) = \mathcal{P}_{\Omega}( \mathbf{L} +  \mathbf{E}),
\label{E:PCP}
\end{aligned}
\end{equation}
where $\lVert\cdot\rVert_*$ denotes the nuclear norm,
$\lVert\cdot\rVert_1$ is the matrix $\ell_1$-norm and $\lambda>0$ is a regularizer.
$\Omega$ represents the set of locations corresponding to the observed entries
of $\mathbf{X}$ (i.e., $(i,j)\in\Omega~\text{if}~m_i=m_j=1$). Then,
$\mathcal{P}_{\Omega}(\mathbf{X})$ is defined as the projection of the matrix
$\mathbf{X}$ on the observed entries $\Omega$, namely
${\mathcal{P}_{\Omega}(\mathbf{X})}_{ij} = x_{ij}~\text{if}~(i,j)\in\Omega$
and ${\mathcal{P}_{\Omega}(\mathbf{X})}_{ij}=0$ otherwise.
The unique solution of the convex optimization problem in Eq.~\ref{E:PCP} is found by employing an Alternating Direction Method of Multipliers-based algorithm~\cite{bertsekas2014constrained}.

\begin{figure}[!t]
    \centering
    \includegraphics[width=0.7\linewidth]{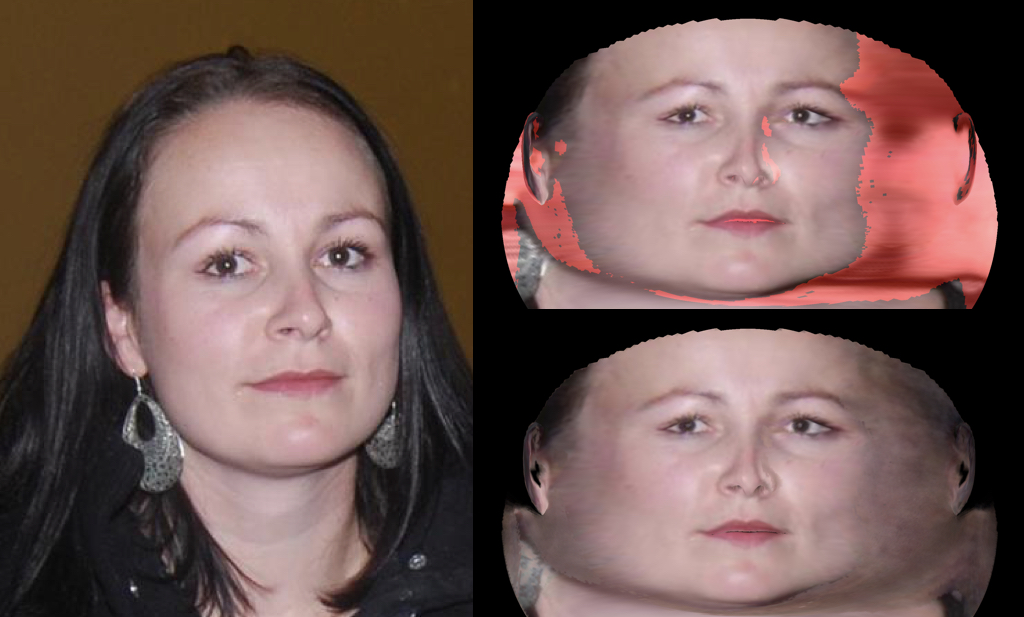}
    \caption{Building an ITW texture model}
\label{fig:itw_extraction}
\end{figure}
The final texture model is created by applying PCA on the set of reconstructed
feature-based textures acquired from the previous procedure. This results in
$\left\lbrace\bar{\mathbf{t}}, \mathbf{U}_t\right\rbrace$,
where $\bar{\mathbf{t}}\in\mathbb{R}^{CN}$ is the mean texture vector and
$\mathbf{U}_t\in\mathbb{R}^{CN\times n_t}$ is the orthonormal basis after
keeping the first $n_t$ principal components. This model can be used to generate
novel 3D feature-based texture instances with the function
$\mathcal{T}: \mathbb{R}^{n_t} \rightarrow \mathbb{R}^{CN}$ as
\begin{equation}
\mathcal{T}(\boldsymbol{\lambda}) = \bar{\mathbf{t}} + \mathbf{U}_t \boldsymbol{\lambda}
\label{equ:texture_instance}
\end{equation}
where $\boldsymbol{\lambda}={[\lambda_1,\ldots,\lambda_{n_t}]}^\mathsf{T}$ are the $n_t$ \emph{texture parameters}.

Finally, an iterative procedure is used in order to refine the texture. That is, we started with the 3D fits provided by using only the 2D landmarks \cite{jourabloo2016large}. Then, a texture model is learned using the above procedure. The texture model was used with the proposed 3DMM fitting algorithm on the same data and texture model was refined.

%% file: fitting.tex
\section{Model Fitting}
\label{sec:fitting}

We propose to fit the 3DMM on an input image using Gauss-Newton iterative
optimization. To this end, herein, we first formulate the cost function and
then present two optimization procedures.

\subsection{Cost Function}
The overall cost function of the proposed 3DMM formulation consists of a
texture-based term, an optional error term based on sparse 2D landmarks and
optional regularization terms on the parameters.

\textbf{Texture reconstruction cost.} The main term of the optimization problem
is the one that aims to estimate the shape, texture and camera parameters that
minimize the ${\ell_2}^2$ norm of the difference between the image feature-based texture that corresponds to the projected 2D locations of the 3D shape instance and the texture instance of the 3DMM. 
Let us denote by $\mathbf{F}=\mathcal{F}(\mathbf{I})$ the feature-based representation with $C$ channels of an input image $\mathbf{I}$ using Eq.~\ref{equ:features}. 
Then, the texture reconstruction cost is expressed as
\begin{equation}
\argmin_{\mathbf{p}, \mathbf{c}, \boldsymbol{\lambda}}
\left\lVert\mathbf{F}\left(\mathcal{W}(\mathbf{p},\mathbf{c})\right) - \mathcal{T}(\boldsymbol{\lambda})\right\rVert^2
\label{equ:data_cost}
\end{equation}
Note that $\mathbf{F}\left(\mathcal{W}(\mathbf{p},\mathbf{c})\right)\in\mathbb{R}^{CN}$ denotes
the operation of sampling the feature-based input image on the projected 2D locations of the 3D shape instance acquired by the camera model (Eq.~\ref{equ:camera_model}).

\textbf{Regularization.} In order to avoid over-fitting effects, we augment the
cost function with two optional regularization terms over the shape and texture
parameters. Let us denote as $\mathbf{\Sigma}_s\in\mathbb{R}^{n_s\times n_s}$
and $\mathbf{\Sigma}_t\in\mathbb{R}^{n_t\times n_t}$ the diagonal matrices with
the eigenvalues in their main diagonal for the shape and texture models,
respectively. Based on the PCA nature of the shape and texture models, it is
assumed that their parameters follow normal prior distributions, i.e.
$\mathbf{p}\sim\mathcal{N}(\mathbf{0},\mathbf{\Sigma}_s)$ and
$\boldsymbol{\lambda}\sim\mathcal{N}(\mathbf{0},\mathbf{\Sigma}_t)$.
We formulate the regularization terms as the ${\ell_2}^2$ of the
parameters' vectors weighted with the corresponding inverse eigenvalues, i.e.
\begin{equation}
\argmin_{\mathbf{p}, \boldsymbol{\lambda}}~c_s\left\lVert\mathbf{p}\right\rVert^2_{\mathbf{\Sigma}_s^{-1}} + c_t\left\lVert\boldsymbol{\lambda}\right\rVert^2_{\mathbf{\Sigma}_t^{-1}}
\label{equ:priors}
\end{equation}
where $c_s$ and $c_t$ are constants that weight the contribution of the regularization
terms in the cost function.

\textbf{2D landmarks cost.} In order to rapidly adapt the camera parameters in
the cost of Eq.~\ref{equ:data_cost}, we further expand the optimization problem
with the term
\begin{equation}
\argmin_{\mathbf{p},\mathbf{c}}~c_l\left\lVert\mathcal{W}_l(\mathbf{p},\mathbf{c}) - \mathbf{s}_l\right\rVert^2
\label{equ:landmarks_cost}
\end{equation}
where $\mathbf{s}_l={\left[x_1,y_1,\ldots,x_L,y_L\right]}^\mathsf{T}$ denotes a
set of $L$ sparse 2D landmark points ($L\ll N$) defined on the image coordinate
system and $\mathcal{W}_l(\mathbf{p},\mathbf{c})$ returns the $2L\times 1$ vector
of 2D projected locations of these $L$ sparse landmarks. Intuitively, this term
aims to drive the optimization procedure using the selected sparse landmarks as
anchors for which we have the optimal locations $\mathbf{s}_l$.
This optional landmarks-based cost is weighted with the constant $c_l$.

\textbf{Overall cost function.} The overall 3DMM cost function is formulated as
the sum of the terms in Eqs.~\ref{equ:data_cost},~\ref{equ:priors},~\ref{equ:landmarks_cost},
i.e.
\begin{equation}
\begin{aligned}
\argmin_{\mathbf{p},\mathbf{c},\boldsymbol{\lambda}}~& \left\lVert\mathbf{F}\left(\mathcal{W}(\mathbf{p},\mathbf{c})\right) - \mathcal{T}(\boldsymbol{\lambda})\right\rVert^2 + c_l\left\lVert\mathcal{W}_l(\mathbf{p},\mathbf{c}) - \mathbf{s}_l\right\rVert^2 +\\
& + c_s\left\lVert\mathbf{p}\right\rVert^2_{\mathbf{\Sigma}_s^{-1}} + c_t\left\lVert\boldsymbol{\lambda}\right\rVert^2_{\mathbf{\Sigma}_t^{-1}}
\label{equ:cost}
\end{aligned}
\end{equation}
The landmarks term as well as the regularization terms are optional and aim to
facilitate the optimization procedure in order to converge faster and to a better
minimum. Note that thanks to the proposed ``in-the-wild'' feature-based texture
model, the cost function does not include any parametric illumination model
similar to the ones in the relative literature~\cite{blanz1999morphable,blanz2003face}, which greatly
simplifies the optimization.

\subsection{Gauss-Newton Optimization}
Inspired by the extensive literature in Lucas-Kanade 2D image
alignment~\cite{baker2004lucas,matthews2004active,papandreou2008adaptive,tzimiropoulos2013optimization,antonakos2015feature,alabort2016unified},
we formulate a Gauss-Newton optimization framework. Specifically, given that
the camera projection model is applied on the image part of
Eq.~\ref{equ:cost}, the proposed optimization has a ``forward'' nature.

\textbf{Parameters update.} The shape, texture and camera parameters are updated
in an additive manner, i.e.
\begin{equation}
\mathbf{p}\leftarrow\mathbf{p}+\Delta\mathbf{p},~~\boldsymbol{\lambda}\leftarrow\boldsymbol{\lambda}+\Delta\boldsymbol{\lambda},~~\mathbf{c}\leftarrow\mathbf{c}+\Delta\mathbf{c}
\end{equation}
where $\Delta\mathbf{p}$, $\Delta\boldsymbol{\lambda}$ and $\Delta\mathbf{c}$
are their increments estimated at each fitting iteration. 
Note that in the case of the quaternion
used to parametrize the 3D rotation matrix, the update is performed as the
multiplication
\begin{equation}
\begin{aligned}
\mathbf{q}\leftarrow & (\Delta\mathbf{q})\mathbf{q} =
\left[\begin{array}{c}\Delta q_0\\\Delta\mathbf{q}_{1:3}\end{array}\right]
\left[\begin{array}{c}q_0\\\mathbf{q}_{1:3}\end{array}\right] = \\
& = \left[\begin{array}{c}\Delta q_0 q_0 - \Delta\mathbf{q}_{1:3}^\mathsf{T}\mathbf{q}_{1:3}\\ \Delta q_0\mathbf{q}_{1:3} + q_0\Delta\mathbf{q}_{1:3}+\Delta\mathbf{q}_{1:3}\times\mathbf{q}_{1:3}\end{array}\right]
\end{aligned}
\end{equation}
However, we will still denote it as an addition for simplicity. Finally, we found that it is beneficial to keep the focal length constant in most cases, due to its ambiguity with $t_z$.

\textbf{Linearization.} By introducing the additive incremental updates on the
parameters of Eq.~\ref{equ:cost}, the cost function is expressed as
\begin{equation}
\begin{aligned}
\argmin_{\Delta\mathbf{p},\Delta\mathbf{c},\Delta\boldsymbol{\lambda}}~& \left\lVert\mathbf{F}\left(\mathcal{W}(\mathbf{p}+\Delta\mathbf{p},\mathbf{c}+\Delta\mathbf{c})\right) - \mathcal{T}(\boldsymbol{\lambda}+\Delta\boldsymbol{\lambda})\right\rVert^2 + \\
& + c_l\left\lVert\mathcal{W}_l(\mathbf{p}+\Delta\mathbf{p},\mathbf{c}+\Delta\mathbf{c}) - \mathbf{s}_l\right\rVert^2 +\\
& + c_s\left\lVert\mathbf{p}+\Delta\mathbf{p}\right\rVert^2_{\mathbf{\Sigma}_s^{-1}} + c_t\left\lVert\boldsymbol{\lambda}+\Delta\boldsymbol{\lambda}\right\rVert^2_{\mathbf{\Sigma}_t^{-1}}
\end{aligned}
\label{equ:cost_with_deltas}
\end{equation}
Note that the texture reconstruction
and landmarks constraint terms of this cost function are non-linear due
to the camera model operation. We need to linearize them around
$(\mathbf{p}, \mathbf{c})$ using first order Taylor series expansion at
$(\mathbf{p}+\Delta\mathbf{p},\mathbf{c}+\Delta\mathbf{c})=(\mathbf{p},\mathbf{c})\Rightarrow(\Delta\mathbf{p},\Delta\mathbf{c})=\mathbf{0}$.
The linearization for the image term gives
\begin{equation}
\begin{aligned}
\mathbf{F}\left(\mathcal{W}(\mathbf{p}+\Delta\mathbf{p},\mathbf{c}+\Delta\mathbf{c})\right) \approx & \mathbf{F}\left(\mathcal{W}(\mathbf{p},\mathbf{c})\right) + \\
& + \mathbf{J}_{\mathbf{F},\mathbf{p}}\Delta\mathbf{p} + \mathbf{J}_{\mathbf{F},\mathbf{c}}\Delta\mathbf{c}
\end{aligned}
\label{equ:image_linearization}
\end{equation}
where $\mathbf{J}_{\mathbf{F},\mathbf{p}}=\nabla\mathbf{F}\left.\frac{\partial\mathcal{W}}{\partial\mathbf{p}}\right|_{\mathbf{p}=\mathbf{p}}$
and
$\mathbf{J}_{\mathbf{F},\mathbf{c}}=\nabla\mathbf{F}\left.\frac{\partial\mathcal{W}}{\partial\mathbf{c}}\right|_{\mathbf{c}=\mathbf{c}}$
are the \emph{image Jacobians} with respect to the shape and camera parameters, respectively. Note that most dense feature-extraction functions $\mathcal{F}(\cdot)$ are non-differentiable, thus we simply compute the gradient of the multi-channel feature image $\nabla\mathbf{F}$.
Similarly, the linearization on the sparse landmarks projection term gives
\begin{equation}
\mathcal{W}_l(\mathbf{p}+\Delta\mathbf{p},\mathbf{c}+\Delta\mathbf{c})\approx\mathcal{W}_l(\mathbf{p},\mathbf{c}) + \mathbf{J}_{\mathcal{W}_l,\mathbf{p}}\Delta\mathbf{p} + \mathbf{J}_{\mathcal{W}_l,\mathbf{c}}\Delta\mathbf{c}
\label{equ:landmarks_linearization}
\end{equation}
where
$\mathbf{J}_{\mathcal{W}_l,\mathbf{p}}=\left.\frac{\partial\mathcal{W}_l}{\partial\mathbf{p}}\right|_{\mathbf{p}=\mathbf{p}}$
and
$\mathbf{J}_{\mathcal{W}_l,\mathbf{c}}=\left.\frac{\partial\mathcal{W}_l}{\partial\mathbf{c}}\right|_{\mathbf{c}=\mathbf{c}}$
are the \emph{camera Jacobians}. Please refer to the supplementary material for more details on the computation of these derivatives.

\subsubsection{Simultaneous}
Herein, we aim to simultaneously solve for all parameters' increments.
By substituting Eqs.~\ref{equ:image_linearization} and~\ref{equ:landmarks_linearization}
in Eq.~\ref{equ:cost_with_deltas} we get
\begin{equation}
\begin{aligned}
&\argmin_{\Delta\mathbf{p},\Delta\mathbf{c},\Delta\boldsymbol{\lambda}}\\
&\left\lVert\mathbf{F}\left(\mathcal{W}(\mathbf{p},\mathbf{c})\right) + \mathbf{J}_{\mathbf{F},\mathbf{p}}\Delta\mathbf{p} + \mathbf{J}_{\mathbf{F},\mathbf{c}}\Delta\mathbf{c} - \mathcal{T}(\boldsymbol{\lambda}+\Delta\boldsymbol{\lambda})\right\rVert^2 + \\
&+ c_l\left\lVert\mathcal{W}_l(\mathbf{p},\mathbf{c}) + \mathbf{J}_{\mathcal{W}_l,\mathbf{p}}\Delta\mathbf{p} + \mathbf{J}_{\mathcal{W}_l,\mathbf{c}}\Delta\mathbf{c} - \mathbf{s}_l\right\rVert^2 +\\
&+ c_s\left\lVert\mathbf{p}+\Delta\mathbf{p}\right\rVert^2_{\mathbf{\Sigma}_s^{-1}} + c_t\left\lVert\boldsymbol{\lambda}+\Delta\boldsymbol{\lambda}\right\rVert^2_{\mathbf{\Sigma}_t^{-1}}
\end{aligned}
\label{equ:simultaneous_cost}
\end{equation}
Let us concatenate the parameters and their increments as
$\mathbf{b}={[\mathbf{p}^\mathsf{T}, \mathbf{c}^\mathsf{T}, \boldsymbol{\lambda}^\mathsf{T}]}^\mathsf{T}$
and
$\Delta\mathbf{b}={[\Delta\mathbf{p}^\mathsf{T}, \Delta\mathbf{c}^\mathsf{T}, \Delta\boldsymbol{\lambda}^\mathsf{T}]}^\mathsf{T}$.
By taking the derivative of the final linearized cost function with respect to
$\Delta\mathbf{b}$ and equalizing with zero, we get the solution
\begin{equation}
\mathbf{b} = -\mathbf{H}^{-1} \left(\mathbf{J}_{\mathbf{F}}^\mathsf{T}\mathbf{e}_{\mathbf{F}} + c_l\mathbf{J}_{\mathcal{W}_l}^\mathsf{T}\mathbf{e}_l + c_s\mathbf{\Sigma}_s^{-1}\mathbf{p} + c_t\mathbf{\Sigma}_t^{-1}\boldsymbol{\lambda}\right)
\end{equation}
where
$\mathbf{H}={\mathbf{J}_{\mathbf{F}}}^\mathsf{T}\mathbf{J}_{\mathbf{F}} + c_l{\mathbf{J}_{\mathcal{W}_l}}^\mathsf{T}\mathbf{J}_{\mathcal{W}_l} + c_s\mathbf{\Sigma}_s^{-1} + c_t\mathbf{\Sigma}_t^{-1}$
is the Hessian with
\begin{equation}
\begin{aligned}
\mathbf{J}_{\mathbf{F}} = &~{\left[\mathbf{J}_{\mathbf{F},\mathbf{p}}^\mathsf{T}, \mathbf{J}_{\mathbf{F},\mathbf{c}}^\mathsf{T}, -\mathbf{U}_t^\mathsf{T}\right]}^\mathsf{T}\\
\mathbf{J}_{\mathcal{W}_l} = &~{\left[\mathbf{J}_{\mathcal{W}_l,\mathbf{p}}^\mathsf{T}, \mathbf{J}_{\mathcal{W}_l,\mathbf{c}}^\mathsf{T}, \mathbf{0}_{n_t\times 2L}\right]}^\mathsf{T}
\end{aligned}
\end{equation}
and
\begin{equation}
\begin{aligned}
\mathbf{e}_{\mathbf{F}} = &~\mathbf{F}\left(\mathcal{W}(\mathbf{p},\mathbf{c})\right) - \mathcal{T}(\boldsymbol{\lambda})\\
\mathbf{e}_l = &~\mathcal{W}_l(\mathbf{p},\mathbf{c}) - \mathbf{s}_l
\end{aligned}
\end{equation}
are the residual terms. The computational complexity of the Simultaneous algorithm per iteration is dominated by the texture reconstruction term as
$\mathcal{O}((n_s+n_c+n_t)^3 + CN(n_s+n_c+n_t)^2)$, which in practice is too slow.

\subsubsection{Project-Out}
We propose to use a Project-Out optimization approach that is much faster than the Simultaneous. The main idea is to optimize on the orthogonal complement of the texture subspace which will eliminate the need to solve for the texture parameters increment at each iteration. By substituting Eqs.~\ref{equ:image_linearization} and~\ref{equ:landmarks_linearization}
into Eq.~\ref{equ:cost_with_deltas} and removing the incremental update on the
texture parameters as well as the texture parameters regularization term,
we end up with the problem
\begin{equation}
\begin{aligned}
\argmin_{\Delta\mathbf{p},\Delta\mathbf{c},\boldsymbol{\lambda}}~& \left\lVert\mathbf{F}\left(\mathcal{W}(\mathbf{p},\mathbf{c})\right) + \mathbf{J}_{\mathbf{F},\mathbf{p}}\Delta\mathbf{p} + \mathbf{J}_{\mathbf{F},\mathbf{c}}\Delta\mathbf{c} - \mathcal{T}(\boldsymbol{\lambda})\right\rVert^2 + \\
& + c_l\left\lVert\mathcal{W}_l(\mathbf{p},\mathbf{c}) + \mathbf{J}_{\mathcal{W}_l,\mathbf{p}}\Delta\mathbf{p} + \mathbf{J}_{\mathcal{W}_l,\mathbf{c}}\Delta\mathbf{c} - \mathbf{s}_l\right\rVert^2 +\\
& + c_s\left\lVert\mathbf{p}+\Delta\mathbf{p}\right\rVert^2_{\mathbf{\Sigma}_s^{-1}}
\end{aligned}
\label{equ:project_out_cost}
\end{equation}
The solution of Eq.~\ref{equ:project_out_cost} with respect to $\boldsymbol{\lambda}$ is
readily given by
\begin{equation}
\boldsymbol{\lambda} = {\mathbf{U}_t}^\mathsf{T} \left(\mathbf{F}(\mathcal{W}(\mathbf{p},\mathbf{c})) + \mathbf{J}_{\mathbf{F},\mathbf{p}}\Delta\mathbf{p} + \mathbf{J}_{\mathbf{F},\mathbf{c}}\Delta\mathbf{c} - \bar{\mathbf{t}}\right)
\label{equ:lambda_solution}
\end{equation}
By plugging Eq.~\ref{equ:lambda_solution} into Eq.~\ref{equ:project_out_cost},
we get
\begin{equation}
\begin{aligned}
\argmin_{\Delta\mathbf{p},\Delta\mathbf{c}}~& \left\lVert\mathbf{F}\left(\mathcal{W}(\mathbf{p},\mathbf{c})\right) + \mathbf{J}_{\mathbf{F},\mathbf{p}}\Delta\mathbf{p} + \mathbf{J}_{\mathbf{F},\mathbf{c}}\Delta\mathbf{c} - \bar{\mathbf{t}}\right\rVert^2_\mathbf{P} + \\
& + c_l\left\lVert\mathcal{W}_l(\mathbf{p},\mathbf{c}) + \mathbf{J}_{\mathcal{W}_l,\mathbf{p}}\Delta\mathbf{p} + \mathbf{J}_{\mathcal{W}_l,\mathbf{c}}\Delta\mathbf{c} - \mathbf{s}_l\right\rVert^2 +\\
& + c_s\left\lVert\mathbf{p}+\Delta\mathbf{p}\right\rVert^2_{\mathbf{\Sigma}_s^{-1}}
\end{aligned}
\label{equ:project_out_cost_2}
\end{equation}
where $\mathbf{P}=\mathbf{E}-\mathbf{U}_t{\mathbf{U}_t}^\mathsf{T}$ is the
orthogonal complement of the texture subspace that functions as the
``project-out'' operator with $\mathbf{E}$ denoting the $CN \times CN$ unitary matrix.
Note that in order to derive Eq.~\ref{equ:project_out_cost_2}, we use the properties
$\mathbf{P}^\mathsf{T}=\mathbf{P}$
and
$\mathbf{P}^\mathsf{T}\mathbf{P}=\mathbf{P}$. By differentiating
Eq.~\ref{equ:project_out_cost_2} and equalizing to zero, we get the solution
\begin{equation}
\begin{aligned}
\Delta\mathbf{p} = &~{\mathbf{H}_{\mathbf{p}}}^{-1} \left( \mathbf{J}_{\mathbf{F},\mathbf{p}}^\mathsf{T} \mathbf{P} \mathbf{e}_{\mathbf{F}} + c_l\mathbf{J}_{\mathcal{W}_l,\mathbf{p}}^\mathsf{T}\mathbf{e}_l + c_s\mathbf{\Sigma}^{-1}_s\mathbf{p}\right)\\
\Delta\mathbf{c} = &~{\mathbf{H}_{\mathbf{c}}}^{-1} \left( \mathbf{J}_{\mathbf{F},\mathbf{c}}^\mathsf{T} \mathbf{P} \mathbf{e}_{\mathbf{F}} + c_l\mathbf{J}_{\mathcal{W}_l,\mathbf{c}}^\mathsf{T}\mathbf{e}_l\right)
\end{aligned}
\label{equ:project_out_solution}
\end{equation}
where
\begin{equation}
\begin{aligned}
\mathbf{H}_{\mathbf{p}} = &~\mathbf{J}_{\mathbf{F},\mathbf{p}}^\mathsf{T}\mathbf{P}\mathbf{J}_{\mathbf{F},\mathbf{p}} + c_l\mathbf{J}_{\mathcal{W}_l,\mathbf{p}}^\mathsf{T}\mathbf{J}_{\mathcal{W}_l,\mathbf{p}} + c_s\mathbf{\Sigma}^{-1}\\
\mathbf{H}_{\mathbf{c}} = &~\mathbf{J}_{\mathbf{F},\mathbf{c}}^\mathsf{T}\mathbf{P}\mathbf{J}_{\mathbf{F},\mathbf{c}} + c_l\mathbf{J}_{\mathcal{W}_l,\mathbf{c}}^\mathsf{T}\mathbf{J}_{\mathcal{W}_l,\mathbf{c}}
\end{aligned}
\end{equation}
are the Hessian matrices and
\begin{equation}
\begin{aligned}
\mathbf{e}_{\mathbf{F}} = &~\mathbf{F}\left(\mathcal{W}(\mathbf{p},\mathbf{c})\right) - \bar{\mathbf{t}}\\
\mathbf{e}_l = &~\mathcal{W}_l(\mathbf{p},\mathbf{c}) - \mathbf{s}_l
\end{aligned}
\end{equation}
are the residual terms. The texture parameters can be estimated at the end of the iterative procedure using Eq.~\ref{equ:lambda_solution}. 

Note that the most expensive operation is $\mathbf{J}_{\mathbf{F},\mathbf{p}}^\mathsf{T}\mathbf{P}$. However, if we first do $\mathbf{J}_{\mathbf{F},\mathbf{p}}^\mathsf{T}\mathbf{U}_t$ and then multiply this result with $\mathbf{U}_t^\mathsf{T}$, the total cost becomes $\mathcal{O}(CNn_tn_s)$. The same stands for $\mathbf{J}_{\mathbf{F},\mathbf{c}}^\mathsf{T}\mathbf{P}$. Consequently, the cost per iteration is 
$\mathcal{O}((n_s+n_c)^3 + CNn_t(n_s+n_c) + CN(n_s+n_c)^2)$
which is much faster than the Simultaneous algorithm.

\textbf{Residual masking.}
In practice, we apply a mask on the texture reconstruction residual of the Gauss-Newton optimization, in order to speed-up the 3DMM fitting. This mask is constructed by first acquiring the set of visible vertexes using z-buffering and then randomly selecting $K$ of them. By keeping the number of vertexes small ($K\approx 5000 \ll N$), we manage to greatly speed-up the fitting process without any accuracy penalty.

%% file: dataset.tex
\section{KF-ITW Dataset}
\label{sec:dataset}

For the evaluation of the 3DMM, we have constructed KF-ITW, the first dataset of 3D faces captured under relatively unconstrained conditions. The dataset consists of $17$ different subjects recorded under various illumination conditions performing a range of expressions (\emph{neutral}, \emph{happy}, \emph{surprise}).
We employed the KinectFusion~\cite{izadi2011kinectfusion,newcombe2011kinectfusion} framework to acquire a 3D representation of the subjects with a Kinect v1 sensor.

The fused mesh for each subject serves as a 3D face ground-truth in which we can evaluate our algorithm and compare it to other methods.
A voxel grid of size $608^3$ was utilized to get the detailed 3D scans of the faces.
In order to accurately reconstruct the entire surface of the faces, a circular motion scanning pattern was carried out. Each subject was instructed to stay still in a fixed pose during the entire scanning process. The frame rate for every subject was constant to $8$ frames per second. After getting the 3D scans from the KinectFusion framework we fit our shape model in a non-rigid manner to get a clear mesh with a distinct number of vertexes for the evaluation process. Finally, each mesh was manually annotated with the iBUG 49 sparse landmark set.

%% file: experiments.tex
\section{Experiments}
\label{sec:experiments}

To train our model, which we label as \textit{ITW}, we use a variant of the Basel Face Model (BFM)~\cite{paysan20093d} that we trained to contain both identities drawn from the original BFM model along with expressions provided by~\cite{cao2014facewarehouse}. 
We trained the ``in-the-wild'' texture model on the images of iBUG, LFPW \& AFW datasets~\cite{sagonas2016faces} as described in Sec.~\ref{sec:texture-model} using the 3D shape fits provided by~\cite{Zhu_2016_CVPR}. Additionally, we elect to use the project-out formulation for the throughout our experiments due its superior run-time performance and equivalent fitting performance to the simultaneous one.

\subsection{3D Shape Recovery}
\label{sec:experiments-quantitative-shape}
Herein, we evaluate our ``in-the-wild'' 3DMM (\textit{ITW}) in terms of 3D shape estimation accuracy against two popular state-of-the-art alternative 3DMM formulations. The first one is a classic 3DMM with the original Basel laboratory texture model and full lighting equation which we term
\textit{Classic}. The second is the texture-less linear model proposed in~\cite{huber2015fitting,huber2016multiresolution} which we refer to as~\textit{Linear}. For \textit{Linear} code we use the Surrey Model with related blendshapes along with the implementation given in~\cite{huber2016multiresolution}. 

\begin{figure}[!t]
    \centering
    \includegraphics[width=\linewidth]{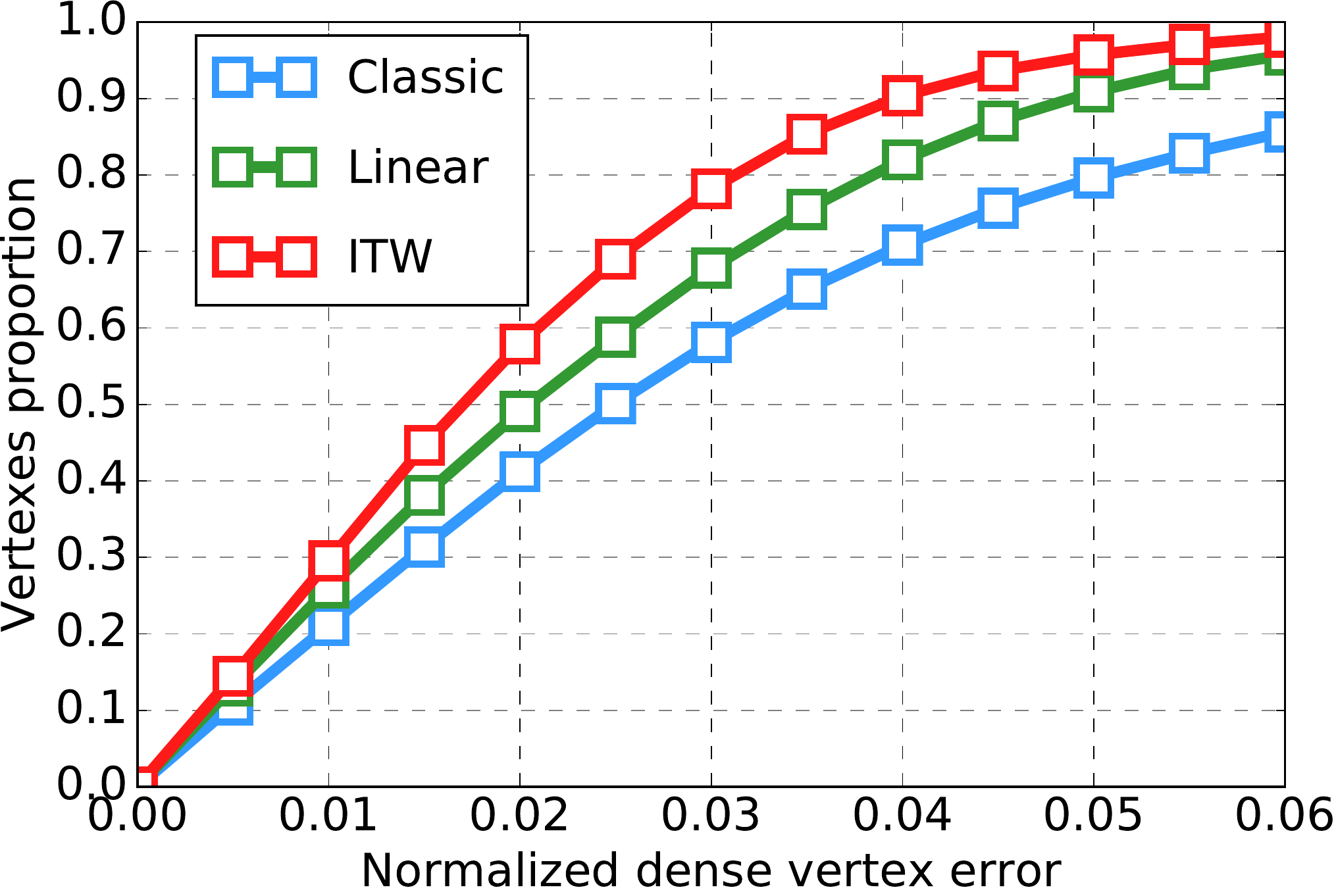}
    \caption{Accuracy results for facial shape estimation on KF-ITW database. The 
    results are presented as Cumulative Error Distributions of the normalized dense vertex error. Table~\ref{tab:kf_dense_fit_error_ced} reports additional measures.}
    \label{fig:kf_dense_fit_error_ced}
\end{figure}

We use the ground-truth annotations provided in the KF-ITW dataset to initialize and fit all three techniques to the ``in-the-wild'' style images in the dataset. The mean mesh of each model under test is landmarked with the same 49-point markup used in the dataset, and is registered against the ground truth mesh by performing a Procrustes alignment using the sparse annotations followed by Non-Rigid Iterative Closest Point (N-ICP) to iteratively deform the two surfaces until they are brought into correspondence.
This provides a per-model `ground-truth' for the 3D shape recovery problem for each image under test.
Our error metric is the per-vertex dense error between the recovered shape and the model-specific corresponded ground-truth fit, normalized by the inter-ocular distance for the test mesh.
Fig.~\ref{fig:kf_dense_fit_error_ced} shows the cumulative error distribution for this experiment for the three models under test. Table~\ref{tab:kf_dense_fit_error_ced} reports the corresponding Area Under the Curve (AUC) and failure rates. The \emph{Classic} model
struggles to fit to the ``in-the-wild'' conditions present in the test set, and performs the worst.
The texture-free \textit{Linear} model does better, but the \textit{ITW} model is most able to recover the facial shapes due to its ideal feature basis for the ``in-the-wild'' conditions.

\begin{figure*}
\centering
\includegraphics[width=0.95\linewidth]{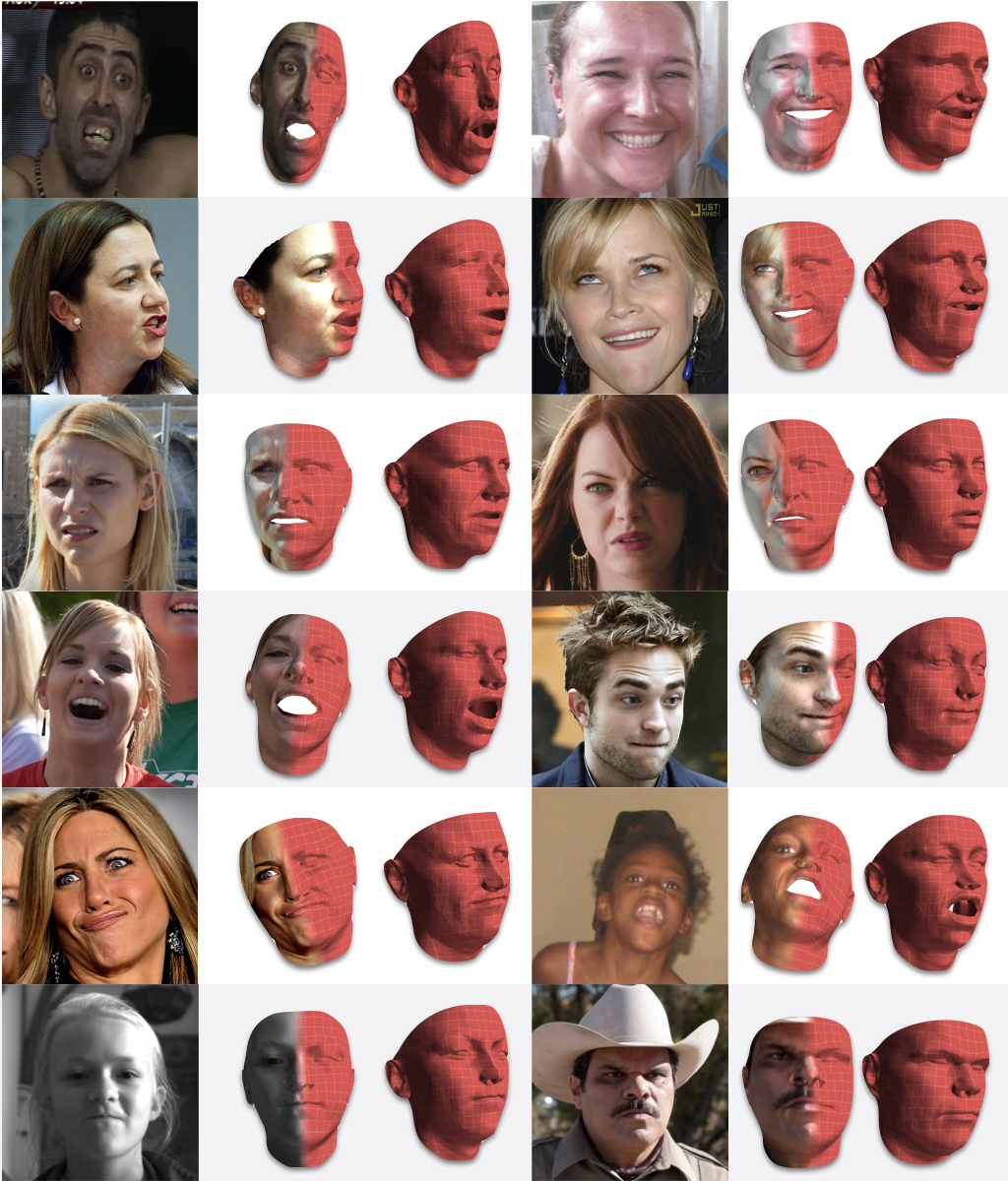}
\caption{Examples of in the wild fits of our \emph{ITW} 3DMM taken from 300W~\cite{sagonas2016faces}.}
\label{fig:example_gallery}
\end{figure*}

Figure~\ref{fig:example_gallery} demonstrates qualitative results on a wide range of fits of ``in-the-wild'' images drawn from the Helen and 300W datasets~\cite{sagonas2016faces,sagonas2013300} that qualitatively highlight the effectiveness of the proposed technique.
We note that in a wide variety of expression, identity, lighting and occlusion conditions our model is able to robustly reconstruct a realistic 3D facial shape that stands up to scrutiny.

\begin{table}[!t]
\centering
\begin{tabular}{|l|cc|}
\hline
\emph{Method} & \emph{AUC} & \emph{Failure Rate (\%)} \\
\hline\hline
\textbf{ITW}     & \textbf{0.678} & \textbf{1.79} \\
Linear  & 0.615 & 4.02 \\
Classic & 0.531 & 13.9 \\
\hline
\end{tabular}
\caption{Accuracy results for facial shape estimation on KF-ITW database. The 
    table reports the Area Under the Curve (AUC) and Failure Rate of the Cumulative Error Distributions of Fig.~\ref{fig:kf_dense_fit_error_ced}.}
\label{tab:kf_dense_fit_error_ced}
\end{table}

\begin{figure}[!t]
    \centering
    \includegraphics[width=\linewidth]{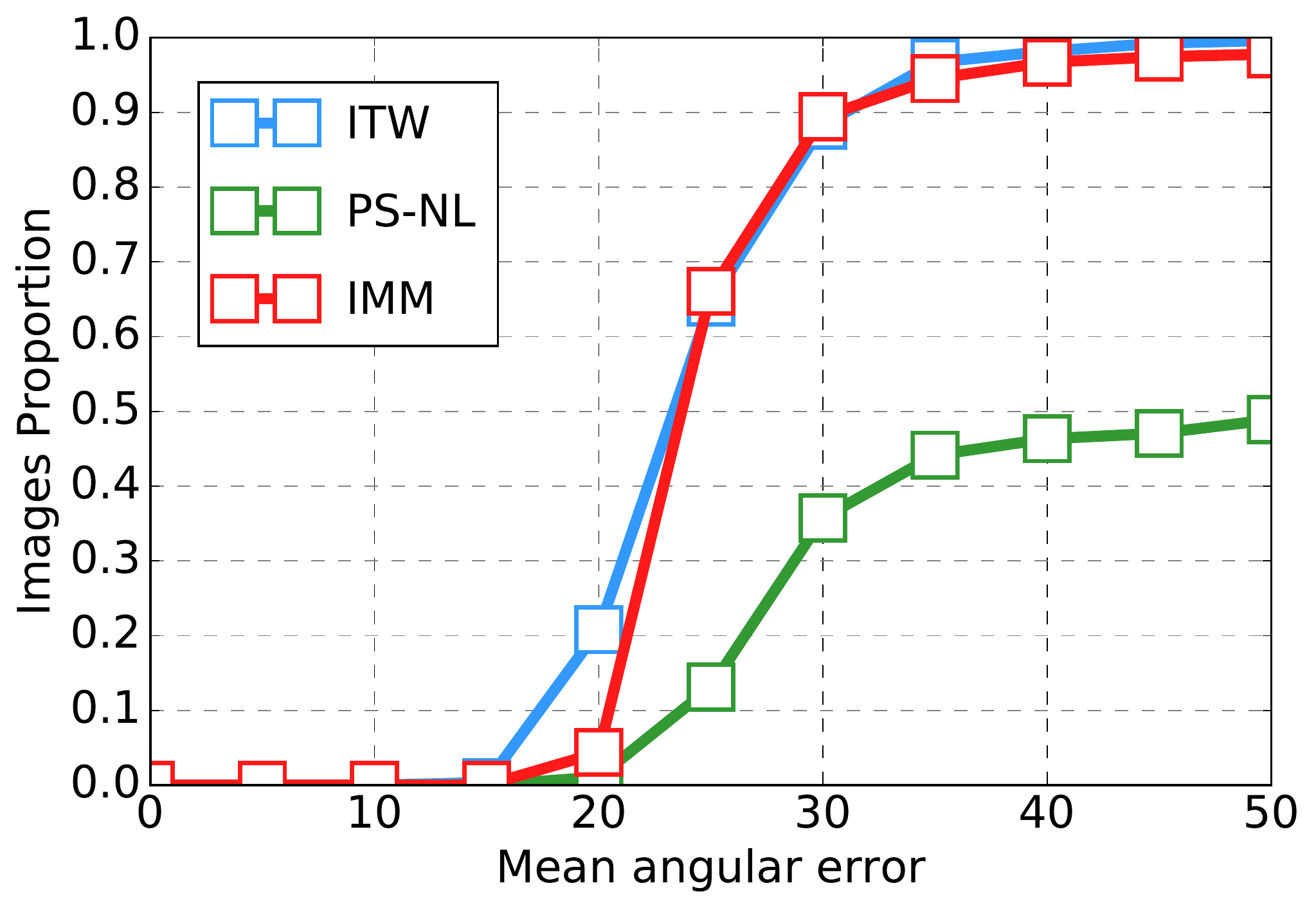}
    \caption{Results on facial surface normal estimation in the form of Cumulative Error Distribution of mean angular error.}
\label{fig:angular_normals_ced}
\end{figure}

\subsection{Quantitative Normal Recovery}
\label{sec:experiments-quantitative-normals}
As a second evaluation, we use our technique to find per-pixel normals and compare against two well established Shape-from-Shading~(SfS) techniques:~\textit{PS-NL}~\cite{basri2007photometric} and \textit{IMM}~\cite{kemelmacher2013internet}. For experimental evaluation we employ images of 100 subjects from the Photoface database~\cite{photoface}. As a set of four illumination conditions are provided for each subject then we can generate ground-truth facial surface normals using calibrated 4-source Photometric Stereo~\cite{marr1978representation}. In~Fig.~\ref{fig:angular_normals_ced} we show the cumulative error distribution in terms of the mean angular error. \textit{ITW} slightly outperforms \textit{IMM} even though both \textit{IMM} and \textit{PS-NL} use all four available images of each subject.

%% file: conclusion.tex
\section{Conclusion}
\label{sec:conclusion}

We have presented a novel formulation of 3DMMs re-imagined for use in ``in-the-wild'' conditions. We capitalise on the annotated ``in-the-wild'' facial databases to propose a methodology for learning an ``in-the-wild'' feature-based texture model suitable for 3DMM fitting without having to optimise for illumination parameters. Furthermore, we propose a novel optimisation procedure for 3DMM fitting. We show that we are able to recover shapes with more detail than is possible using purely landmark-driven approaches.
Our newly introduced ``in-the-wild'' KinectFusion dataset allows for the first time a quantitative evaluation of 3D facial reconstruction techniques in the wild, and on these evaluations we demonstrate that our in the wild formulation is state of the art, outperforming classical 3DMM approaches by a considerable margin.